\documentclass[runningheads]{llncs}
\usepackage{graphicx}
\usepackage{multirow} 
\begin{document}
\title{Transformer-based Automatic Speech Recognition of Formal and Colloquial Czech in MALACH Project}
\titlerunning{Transformer-based ASR of Formal and Colloquial Czech}
%
\author{
Jan Lehečka\inst{1}\orcidID{0000-0002-3889-8069} \and
Josef V. Psutka\inst{1}\orcidID{0000-0003-4761-1645} \and Josef Psutka\inst{1}\orcidID{0000-0002-0764-3207}
}
\authorrunning{J. Lehečka et al.}
%

\institute{Department of Cybernetics, University of West Bohemia in Pilsen, Czech Republic\\
\email{\{jlehecka,psutka\_j,psutka\}@kky.zcu.cz}}
\maketitle              
\begin{abstract}
Czech is a very specific language due to its large differences between the formal and the colloquial form of speech. While the formal (written) form is used mainly in official documents, literature, and public speeches, the colloquial (spoken) form is used widely among people in casual speeches. This gap introduces serious problems for ASR systems, especially when training or evaluating ASR models on datasets containing a lot of colloquial speech, such as the MALACH project. In this paper, we are addressing this problem in the light of a new paradigm in end-to-end ASR systems -- recently introduced self-supervised audio Transformers. Specifically, we are investigating the influence of colloquial speech on the performance of Wav2Vec 2.0 models and their ability to transcribe colloquial speech directly into formal transcripts. We are presenting results with both formal and colloquial forms in the training transcripts, language models, and evaluation transcripts.

\keywords{Wav2Vec 2.0 \and Colloquial speech \and ASR.}
\end{abstract}

\section{Introduction}
Formal Czech differs a lot from the colloquial Czech. Almost 20\% of Czech words have different transcription in both varieties \cite[p.~250]{Tahal}. This gap between the everyday, colloquial language, and the official codified formal language emerged during the Czech National Revival back in the 1830s when a group of Czech writers, poets, translators, editors, and teachers established new grammar rules and vocabularies independent of German influence. They took inspiration from other Slavic languages and outdated Czech Bible texts. However, common people did not adopt these new rules and words into their spoken language creating a very specific widely-spoken vernacular that persists to this day \cite{10.2307/24599659}.

The gap between formal and colloquial Czech constitutes a serious problem for Automatic Speech Recognition (ASR) systems which automatically transcribe -- possibly colloquial -- spoken utterances into formal text \cite{1306515}. The usual way how to deal with this phenomenon in a common Large-Vocabulary Continuous Speech Recognition (LVCSR) system is to train the acoustic model with colloquial phonetic transcripts, define alternative (colloquial) pronunciations for formal words in the lexicon and finally use a formal language model to decode the speech into a formal transcript \cite{Psutka_2005,MALACH2}. 

In the recent few years, self-supervised neural networks became a very popular alternative to LVCSR systems in speech recognition tasks. A significant milestone was the introduction of the Transformer architecture \cite{vaswani2017attention} into ASR systems \cite{baevski2020wav2vec,Baevski2020EffectivenessOS,hsu2021hubert,liu2021tera,chen2021wavlm}.
The most studied transformer-based ASR model
architecture is Wav2Vec 2.0 \cite{baevski2020wav2vec}. It is an end-to-end speech recognizer that alleviates the need for word pronunciation modeling and does not require any alignment of data. It is a single model converting the raw audio signal from the input into the sequence of tokens on the output, no meter whether these tokens are graphemes, phonemes, word pieces, or other speech units. Thus, the model has a very interesting ability: when the input audio data during fine-tuning contain colloquial speech and the target transcripts are in the formal Czech, it could internally learn the mapping between the two forms without any engineering or manual effort. In this paper, we are investigating the extent of this ability of Wav2Vec models.

\section{MALACH Project}
The whole story of the MALACH project began in 1994, when after the premiere of the film “Schindler's List”, many survivors turned to Steven Spielberg to tell him their stories about the Holocaust. Inspired by these requests, Spielberg decided to establish the Shoah Visual History Foundation (VHF) so that as many survivors as possible could record their stories and save them for future generations. Nowadays are these video interviews located in the Shoah Foundation Institute at the University of Southern California (USC-SFI) along with another 54,000 video interviews with witnesses of the history of the entire 20th century.

The Shoah part of the archive contains testimonies in 32 languages of personal memories of survivors of the World War II Holocaust, in total it is 116,000 hours of video. Interviews (in all languages) contain natural, unrestricted speech, full of disfluencies, emotional excitements, heavy accents, and are often influenced by the high age of speakers (problems with keeping ideas). More than 550 testimonies are in the Czech (almost 1000 hours of video).

In 2014, the Linguistic Data Consortium (LDC) released the Czech part of the MALACH project \cite{MALACHcz}. There were published 420 testimonies along with their transcripts. The release contains 400 randomly selected testimonies for the purpose of acoustic model training. As only 15-minute segments were transcribed for each testimony, the acoustic training part, therefore, consists of 100 hours of Czech speech from theoretically up to 800 speakers (interviewer and interviewee). The rest of the Czech MALACH corpus consists of 20 testimonies, which have been completely transcribed and are intended for development (10 testimonies, i.e. 20 speakers) and testing (10 testimonies, i.e. 20 speakers) purposes. (see Tab.~\ref{tab:stats} for details).

\begin{table}[t]
  \caption{Statistics of training and test data-sets of the Czech part of the MALACH project.}
  \label{tab:stats}
  \centering
	  \begin{tabular}{lcc}
	  \hline
    \multicolumn{1}{c}{}& ~~~Train~~~ & ~~~Test~~~ \\
     \cline{2-3}
    ~~~\# of speakers             ~~~	& 776   & 20    \\
    ~~~\# of words                ~~~	& 49k   & 10.3k \\
	~~~\# of tokens               ~~~	& 715k  & 63k   \\
	~~~dataset length $[$hours$]$ ~~~	& 87.5  & 8.9   \\
	\hline
  \end{tabular} 
\end{table}

\section{Formal vs. Colloquial Czech}
\label{sec:formrules}
During the annotation process of the Czech Malach corpus, the transcribers were instructed to use the orthographic transcription of colloquial words (i.e., not to “formalize” them artificially) to bring the transcripts as close as possible to what was actually said. There were several reasons for this decision. Firstly, this procedure was very beneficial for classical acoustic modeling, because the resulting transcription is very close to the actual phonetic realization of the word. Secondly, transcribing colloquial sentences using formal words is not an easy task, especially for transcribers without a solid linguistic background. Another problem solved by the colloquial method of transcription was no need to unify the transcription of foreign words.

On the other hand, the effect of the abundance of colloquial words on the language model is rather negative. The orthographic transcription of colloquial words causes an unnecessary growth of the lexicon. There are often several different colloquial variants corresponding to one formal word form. Consequently, the already sparse language model training data became even sparser. To take advantage of formal word forms in language modeling, we decided to “formalize” the lexicon. We went through a lexicon built from the original (orthographic) transcriptions and added a corresponding standard form to each colloquial word form, but only in cases where it was unambiguous. The normalization of manual transcripts not only made the parameters of the estimated language model more robust but also brought this main and most useful source for language modeling much closer to other potential formal text sources. More details on this process can be found in \cite{PsutkaJ_2004_Issuesinannotation}.

A good example of the ambiguity of such a formalization is the word \emph{sem}. While in formal Czech this word means \emph{sem (here)}, in colloquial Czech it is also used instead of the correct form \emph{jsem ((I) am)} which naturally occurs quite frequently (the fourth most frequent word in the corpus). To distinguish which formal variant of a word \emph{sem} is the correct one, we would have to use a larger word context or better use a sophisticated method of text understanding. Nevertheless, by formalizing the lexicon, we found more than 13k unambiguous rules that reduced the number of colloquial words by almost 85\%.

In order to illustrate that the number of colloquial forms for a single formal word form can be really high, we present a fragment from the “formalized” lexicon in Tab.~\ref{tab:formal}. The new “formalized” text corpus was created by automatically replacing colloquial words in the original transcripts with their formal counterparts using the above-mentioned 2-column lexicon. Note that such a procedure does not take into account the word context, and therefore the formalization process is far from perfect.

\begin{table}[t]
  \caption{Example of formalization rules.}
  \label{tab:formal}
  \centering
	  \begin{tabular}{ccc}
	  \hline
		formal  & ~colloquial~ & ~ \emph{in English}\\
		\hline
		\hline
		\multirow{2}{*}{~~~odjet~~~} & ~ odejet odjec odject vodjet & \multirow{2}{*}{~~~\emph{to leave}~~~}\\
		              &  vodejet vodject vodeject\\
    \hline				
		\multirow{2}{*}{~~~odtamtud~~~} & ~	odtamta\v{d} odtamtu\v{d} vodtamta\v{d} vodtamtud & \multirow{2}{*}{~~~\emph{from there}~~~}\\
		                                &    vodtamtu\v{d} votamta\v{d} votamtu\v{d}\\				
		\hline
		bývalý & bejvalej bejvalý bývalej	& 	\emph{former}\\
		\hline
  \end{tabular} 
\end{table}

\section{Wav2Vec 2.0}
Wav2Vec 2.0 model \cite{baevski2020wav2vec} is one of the current state-of-the-art models for ASR.
It is a deep neural network pretrained to reconstruct the corrupted signals. The input raw audio signal is processed by a multi-layer convolutional neural network into a sequence of latent-speech representations which are fed into a multi-layer Transformer \cite{vaswani2017attention}. Only the encoder part of the full encoder-decoder Transformer architecture is used.
The output of the Transformer is a sequence of frame-level contextualized speech representations encoding both the frame itself and its context in the signal. This approach is motivated by very successful self-supervised text-based Transformers solving Natural Language Processing (NLP) and Natural Language Understanding (NLU) tasks \cite{devlin2018bert}.

The training of Wav2Vec models consists of two phases: pretraining and fine-tuning. During the first self-supervised pretraining phase, the model learns contextualized speech representations from large-scale unlabeled audio datasets.
This approach is motivated by the learning skills of infants, who do not learn to understand speech by reading its transcripts, but rather by listening to adults around them and trying to catch the meaning from the context.
By masking latent representations of the raw waveform and solving a contrastive task over quantized speech representations, the model learns contextualized representations jointly with discrete speech units without the need for any annotations or labels.

Since labeled data could be very expensive and precious, the pretraining phase equips the model with deep knowledge about the speech signals mined out from tens of thousands of hours of unlabeled speech. This knowledge constitutes a great advantage over models trained from scratch using labeled data only. From this point of view, the pretrained weights of the Wav2Vec model could be seen as very clever initializations of the model weights for supervised training.

During the second supervised fine-tuning phase,
the model transfers the pretrained knowledge into the ASR task. For input speech signals, the speech representations are fed into Connectionist Temporal Classification (CTC) layer \cite{graves2006connectionist} and the most probable sequences of graphemes are decoded. The model is fine-tuned with frozen feature-encoder weights from labeled data optimizing the CTC loss.

CTC is an alignment-free method for grouping audio frames belonging to the same output token in order to convert a sequence of speech representations (one per audio frame) into a much shorter sequence of output tokens.
The CTC classification process can be described -- in a simplified way -- in 3 steps: (1) assign the most probable output token to each audio frame, (2) group sequences with the same tokens into a single token, and (3) remove blank tokens. Tokens are usually graphemes (i.e. characters including also a word delimiter) but could be any speech units.

\section{Experimental Setup}

\subsection{Pretraining}
Public monolingual Wav2Vec models for non-English languages are very rare. For the Czech language, there are none. 
However, there are several public multilingual pretrained models of sizes from large \cite{conneau2020unsupervised} to extremely large \cite{babu2021xls}. These models included also Czech in the pretraining datasets. 
The common practice with these models is to select the most suitable pretrained model and fine-tune it on the labeled ASR data from the target language. 
Since we were not satisfied with results from multilingual models and, at the same time, we had access to large unlabeled datasets and a high-performance GPU cluster, we decided to pretrain our own base-sized monolingual Wav2Vec model from scratch and released it to the public.

Self-supervised audio transformers are known to scale well with the size of pretraining data, even with extremely huge datasets \cite{babu2021xls}. Hence, we tried to gather as much public and in-house unlabeled audio data as possible. Together, we were able to collect more than 80 thousand hours of Czech speech. The collection includes recordings from radio (22k hours), unlabeled data from VoxPopuli dataset \cite{wang-etal-2021-voxpopuli} (18.7k hours), TV shows (15k hours), shadow speakers (12k hours), sports (5k hours), telephone data (2k hours), and a smaller amount of data from several other domains. We included also raw unlabeled audio data from the MALACH project (1k hours).

Since the feature extraction of the input signal is limited by the memory of GPUs in use, we sliced all records not to exceed 30\,s, which we found to be a reasonable input size for batching. 

We followed the same pretraining setup as for the base Wav2Vec 2.0 model in \cite{baevski2020wav2vec}. We pretrained the model for 400 thousand steps with a batch size not exceeding 1.6 hours, corresponding to more than 11 epochs over the dataset. 
The pretraining took about two weeks on a machine with four NVIDIA A100 GPUs. 
We released our pretrained model under the nickname \emph{ClTRUS} (abbreviation for \textbf{C}zech \textbf{l}anguage \textbf{TR}ransformer from \textbf{U}nlabeled \textbf{S}peech) for public non-commercial use\footnote{Available at \url{https://huggingface.co/fav-kky/wav2vec2-base-cs-80k-ClTRUS}}.  We are not aware of any similar model for Czech mentioned in the literature so far.

\subsection{Fine-tuning}
When fine-tuning models, we used the same setup as in \cite{baevski2020wav2vec}, i.e. we trained the pretrained model for 80 thousand update steps with the peak learning rate of $2 \times 10^{-5}$ and the batch size about 27 minutes of audio, resulting in 270 training epochs over the dataset. We removed non-speech events and punctuation from the transcripts and mapped texts into lowercase.
We used implementation from the \texttt{Fairseq} tool\footnote{\url{https://github.com/pytorch/fairseq}} to fine-tune models.

First, we trained the colloquial model, denoted as \texttt{W2V\textsubscript{colloq}}, from the original transcripts. Since annotators were instructed to transcribe the speech in the spoken form, i.e. exactly as it was spoken in the underlying speech, these transcripts are mainly in colloquial Czech.
However, it is in fact a mix of both forms, because some people tend to speak more formally when giving an interview, and sometimes annotators were not able to distinguish between the two forms, especially in the strong emotional and heavily accented speeches. We left the formal words untouched as the rules from formal to colloquial form would be ambiguous.

\begin{figure}[tbh]
\begin{center}
\includegraphics[width=0.8\textwidth]{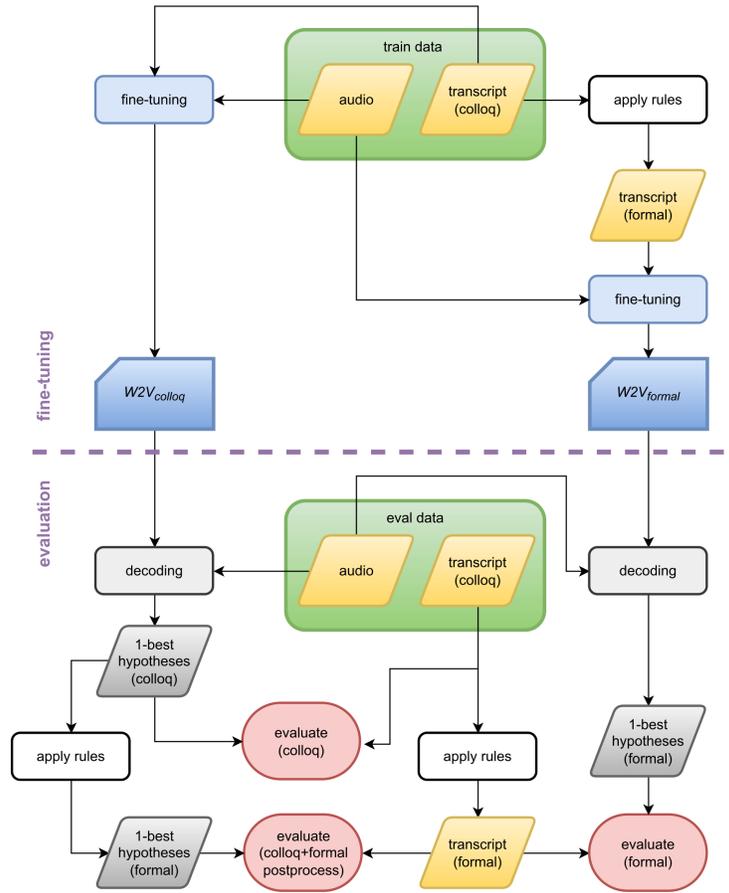}
\caption{Scheme of fine-tuning and evaluation.} \label{fig:schema}
\end{center}
\end{figure}

After that, we transformed the original transcripts into formal Czech using the prepared set of rules (see Sec.~\ref{sec:formrules}) and fine-tuned the second model, denoted as \texttt{W2V\textsubscript{formal}}. The whole fine-tuning process is depicted in the upper part of  Fig.~\ref{fig:schema}. The fine-tuning of each model took about 14 hours on a machine with four NVIDIA A100 GPUs. 

\subsection{Decoding}
\label{sec:decoding}
We studied two different decoding setups: (1) Connectionist Temporal Classification (CTC) \cite{graves2006connectionist}, which is the training loss we used during fine-tuning of the models, and (2) CTC beam search decoder with a Language Model (LM). 
Decoding setup (1) is a grapheme-based lexicon-free speech recognition without any language constraints. The only orthography-related knowledge the model could learn is the training transcripts fed in during the fine-tuning. Wav2Vec with the CTC decoding setup (1) decodes also word delimiters, so it is an end-to-end ASR system, which can be evaluated using standard word-based metrics like word error rate.

Decoding setup (2) incorporates an LM into the CTC beam search decoder which usually improves the speech recognition accuracy by bringing useful language information into the decoding process and penalizing improbable outputs. 
For our experiments, we prepared 2 different word-based n-gram LMs: (a) \texttt{LM\textsubscript{colloq}} trained from all colloquial transcripts, and (b) \texttt{LM\textsubscript{formal}} trained from the formalized training transcripts, i.e. from the training data of \texttt{W2V\textsubscript{formal}} model. We limited the maximum order of models to 4-grams for both LMs.

We used implementation from \texttt{Transformers} \cite{wolf-etal-2020-transformers} for CTC decoding and \texttt{pyctcdecode}\footnote{\url{https://github.com/kensho-technologies/pyctcdecode}} decoder for CTC beam search decoder with n-gram LM. To train LMs, we used \texttt{KenLM} \cite{heafield2011kenlm} and mapped all texts into lowercase.

\subsection{Evaluation}
\label{sec:eval}
Both decoding setups described in Sec.~\ref{sec:decoding} generated a 1-best hypothesis for each input signal. We aligned decoded hypotheses and reference transcripts using the minimum edit distance and evaluated the standard Word Error Rate (WER) and Character Error Rate (CER).

To evaluate colloquial models, we used the original reference transcripts processed in the same way as the training transcripts, i.e. we removed non-speech events and punctuation and mapped texts into lowercase. We denote test dataset with this colloquial reference as \texttt{TEST\textsubscript{colloq}}. To evaluate formal models, we further converted the colloquial reference texts into formal texts using the prepared set of rules (see Sec.~\ref{sec:formrules}) and thus generated the test dataset with formal reference transcripts, denoted as \texttt{TEST\textsubscript{formal}}. 

We evaluated all combinations of formal and colloquial models and LMs against both formal and colloquial reference transcripts. From these combinations, we are particularly interested in three real-world scenarios:
\begin{enumerate}
  \item Evaluation of the colloquial model, i.e. how well \texttt{W2V\textsubscript{colloq}} model with \texttt{LM\textsubscript{colloq}} transcribes the colloquial speech (thus evaluated against \texttt{TEST\textsubscript{colloq}} dataset).
  \item Evaluation of the formal model, i.e. how well \texttt{W2V\textsubscript{formal}} model with \texttt{LM\textsubscript{formal}} transcribes the colloquial speech into formal Czech (thus evaluated against \texttt{TEST\textsubscript{formal}} dataset). This scenario is particularly interesting as it evaluates how well the Wav2Vec model internally learns the mapping between the two forms without any engineering or manual effort.
  \item Transcripts generated from \texttt{W2V\textsubscript{colloq}} with \texttt{LM\textsubscript{colloq}} post-processed by rule-based formalization of texts evaluated against \texttt{TEST\textsubscript{formal}} dataset. 
  This scenario shows how the Wav2Vec model can use data prepared with a great manual effort for a standard LVCSR system in order to generate formal transcripts.
  We denote this colloquial model with Formalization Post-processing as \texttt{W2V\textsubscript{colloq}+FP}
\end{enumerate}

These three scenarios are depicted in a flowchart diagram in the bottom part of Fig.~\ref{fig:schema} and corresponding error rates will be underlined in the results table.

Note, that the numbers of reference words in \texttt{TEST\textsubscript{colloq}} and \texttt{TEST\textsubscript{formal}} differ due to multi-word replacements in the rules. While the formal transcripts consist of 62\,690 words, the colloquial has 62\,918 words, so results evaluated against \texttt{TEST\textsubscript{formal}} and \texttt{TEST\textsubscript{colloq}} are not exactly comparable.

\section{Results}
Results of our experiments are tabulated in Tab.~\ref{tab:results}. First, we evaluated the existing LVCSR system developed specifically for MALACH dataset \cite{MALACH2}. 
The system was a CNN-TDNN LF-MMI with iVectors, sMBR criterion, and system-specific 3-gram LM denoted as \texttt{LM\textsubscript{LVCSR}}. The system was trained to transcribe colloquial speech into formal form, so we report only results evaluated against \texttt{TEST\textsubscript{formal}}. A comparison of this system with the formal Wav2Vec model clearly reveals the superiority of transformer-based ASR systems. 

\setlength{\tabcolsep}{0.5em}
\begin{table}[htb]
\caption{WER [\%] and CER [\%] of colloquial and formal models evaluated against colloquial and formal evaluation datasets (\texttt{TEST\textsubscript{colloq}} and \texttt{TEST\textsubscript{formal}}). 
Each Wav2Vec model was decoded using three different decoding setups: as an end-to-end ASR with no LM and with the beam search CTC decoder with \texttt{LM\textsubscript{formal}} and \texttt{LM\textsubscript{colloq}} (see Sec.~\ref{sec:decoding}). Underlined values correspond to scenarios we are particularly interested in (see Sec.~\ref{sec:eval}). Bold values are the best error rates for each model.}
\label{tab:results}
\begin{center}
\begin{tabular}{lcccccc}
\hline
  &   & \multicolumn{2}{c}{\texttt{TEST\textsubscript{colloq}}} & & \multicolumn{2}{c}{\texttt{TEST\textsubscript{formal}}} \\
  \cline{3-4} \cline{6-7}
\textbf{model} & \textbf{LM} & WER & CER & & WER & CER \\
\hline
LVCSR               & \texttt{LM\textsubscript{LVCSR}} & - & - & &14.71 & 5.25  \\
\hline
\texttt{W2V\textsubscript{colloq}}      & -             & 12.24 & \textbf{3.58} &  &  	19.73 & 5.28 \\
                    & \texttt{LM\textsubscript{formal}} & 13.85 & 4.05 &  & 	15.96 & 4.68 \\
                    & \texttt{LM\textsubscript{colloq}} & \underline{\textbf{11.55}} & \underline{3.64} &  & 	18.99 & 5.27 \\
\hline
\texttt{W2V\textsubscript{formal}}      & -             & 19.17 & 5.07 &  & 	11.52 & 3.32 \\
                    & \texttt{LM\textsubscript{formal}} & 18.60 & 5.19 &  & 	\underline{\textbf{10.48}} & \underline{\textbf{3.31}} \\
                    & \texttt{LM\textsubscript{colloq}} & 18.60 & 5.16 &  & 	10.85 & 3.37 \\
\hline
\texttt{W2V\textsubscript{colloq}+FP} & -             & 19.02 & 5.05 &  & 	11.18 & 3.33 \\
                    & \texttt{LM\textsubscript{formal}} & 18.47 & 5.07 &  & 	11.09 & 3.53 \\
                    & \texttt{LM\textsubscript{colloq}} & 18.47 & 5.12 &  & 	\underline{\textbf{10.43}} & \underline{\textbf{3.30}} \\
\hline
\end{tabular}
\end{center}
\end{table}

As for the Wav2Vec models, the best results evaluated against \texttt{TEST\textsubscript{colloq}} are significantly higher (i.e. worse) than best results evaluated against \texttt{TEST\textsubscript{formal}}. It is mainly because the colloquial Czech does not have codified rules and one formal word could have many possible colloquial forms. Each speaker can use -- based on his or her geographical background -- a different set of colloquial words in the speech. Moreover, each annotator can perceive the spoken colloquial forms differently, especially in the strong emotional and heavily accented speeches. This ambiguity of transcribed speech leads to confusion when training and evaluating the colloquial models.

If we compare the underlined results of the last two Wav2Vec models in Tab.~\ref{tab:results} (corresponding to scenarios 2. and 3. from Sec.~\ref{sec:eval}), we see very similar error rates. The \texttt{W2V\textsubscript{colloq}+FP} is slightly better, which we found to be caused by an occasional incorrect exact match of formalized hypotheses with the formalized reference, as both were generated using the same rules.
After analyzing errors from \texttt{W2V\textsubscript{formal}} model, we found that many recognition errors were actually errors in the reference as the rules were not covering all occurrences of colloquial form in the reference. For example, the formal reference contained (incorrectly) the word “německýho” (colloquial inflected form meaning “German”), because it was not covered by mapping rules due to its non-existence in training transcripts. Formalized output from \texttt{W2V\textsubscript{colloq}+FP} exactly matched the reference for the same reason, so there was no error counted.
\texttt{W2V\textsubscript{formal}} predicted the correct formal form “německého”, which was, however, wrongly counted as a recognition error due to an error in the reference. 
We didn't make more effort to clean the reference transcripts and fix these errors as they were infrequent and it would cost a lot of manual work with only a little effect on the error rates. 
Nevertheless, observing these types of errors was a clear sign of the generalization ability of the \texttt{W2V\textsubscript{formal}} model and we can conclude that \texttt{W2V\textsubscript{formal}} is -- despite slightly higher error rates -- a more useful model than rule-based \texttt{W2V\textsubscript{colloq}+FP} because of its generalization ability.

To sum up the results, Wav2Vec models are significantly better ASR systems for the MALACH project than LVCSR systems. They are able to learn the mapping from colloquial speech into a formal transcript and generalize this skill also to words not observed in training data, which is a more beneficial solution than limited rule-based formalization post-processing of the colloquial model. Moreover, the Wac2Vec's internal mapping from colloquial speech to formal transcripts could make the acquisition of training transcripts much simpler as the annotators could be instructed to transcribe the speech directly into formal Czech alleviating the problems with ambiguous colloquial transcripts and manual listing of rules.

\section{Conclusion}
In this paper, we showed that the new paradigm models in ASR -- Transformer-based models with CTC decoder (specifically Wav2Vec 2.0) -- have a very interesting ability to learn how to transcribe Czech colloquial speech directly into formal transcripts. Such models not only perform better than common LVCSR systems, but also alleviate the need for complicated and ambiguous colloquial annotations, data alignments, phonetic transcriptions, and pronunciation lexicons. When collecting training transcripts for a new ASR dataset, we can instruct annotators just to transcribe the speech directly into formal Czech sentences, which is codified and unambiguous form, and that's all that is needed for the Wav2Vec model to be fine-tuned. From the formal transcript and raw audio signal, the model is able to learn the alignment between the speech signal frames and graphemes, and also how to generalize the conversion between the colloquial speech and formal text. We believe our findings will simplify and accelerate the acquisition of training data for new challenging datasets containing a lot of colloquial speech.

\subsubsection*{Acknowledgments.}
This research was supported by the ITI project of the Ministry
of Education of the Czech Republic CZ.02.1.01/0.0/0.0/17 048/0007267 InteCom.
Computational resources were supplied by the project "e-Infrastruktura CZ" (e-INFRA CZ LM2018140 ) supported by the Ministry of Education, Youth and Sports of the Czech Republic.

\bibliographystyle{splncs04}
\bibliography{tsd1163a}

\end{document}